\def\year{2018}\relax
\documentclass[letterpaper]{article} 
\usepackage{aaai18}  
\usepackage{times}  
\usepackage{helvet}  
\usepackage{courier}  
\usepackage{url}  
\usepackage{graphicx}  

\usepackage{times}
\usepackage{epsfig}
\usepackage{graphicx}
\usepackage{amsmath}
\usepackage{amssymb}

\usepackage{slashbox}
\usepackage{tabularx}
\usepackage[table]{xcolor}

\usepackage{graphicx}
\usepackage{amsmath,amssymb} 
\usepackage{color}

\usepackage{amssymb}

\usepackage{url}

\usepackage{algorithm}               
\usepackage{algorithmic}            
\usepackage{multirow}                
\usepackage{xcolor} 

\begin{document}
%
\title{Predicting Aesthetic Score Distribution \\ through Cumulative Jensen-Shannon Divergence}
\author{Xin Jin$^1$, Le Wu$^1$, Xiaodong Li$^1$, Siyu Chen$^1$, Siwei Peng$^3$,\\ \Large{\textbf{ Jingying Chi$^3$, Shiming Ge$^{4,*}$, Chenggen Song$^2$, Geng Zhao$^1$}}\\
$^1$Department of Computer Science and Technology, 
Beijing Electronic Science and Technology Institute, Beijing, 100070, China
\\
$^2$Department of Information Security, 
Beijing Electronic Science and Technology Institute, Beijing, 100070, China
\\
$^3$College of Information Science and Technology, Beijing University of Chemical Technology, Beijing 100029, China
\\
$^4$Institute of Information Engineering, Chinese Academy of Sciences, Beijing, 100093, China
\\
$*$Corresponding author email: geshiming@iie.ac.cn\\
}
\maketitle
\begin{abstract}
Aesthetic quality prediction is a challenging task in the computer vision community because of the complex interplay with semantic contents and photographic technologies. Recent studies on the powerful deep learning based aesthetic quality assessment usually use a binary high-low label or a numerical score to represent the aesthetic quality. However the scalar representation cannot describe well the underlying varieties of the human perception of aesthetics. In this work, we propose to predict the aesthetic score distribution (i.e., a score distribution vector of the ordinal basic human ratings) using Deep Convolutional Neural Network (DCNN). Conventional DCNNs which aim to minimize the difference between the predicted scalar numbers or vectors and the ground truth cannot be directly used for the ordinal basic rating distribution. Thus, a novel CNN based on the Cumulative distribution with Jensen-Shannon divergence (CJS-CNN) is presented to predict the aesthetic score distribution of human ratings, with a new reliability-sensitive learning method based on the kurtosis of the score distribution, which eliminates the requirement of the original full data of human ratings (without normalization). Experimental results on large scale aesthetic dataset demonstrate the effectiveness of our introduced CJS-CNN in this task.
\end{abstract}

\section{Introduction}
\label{sec:Intro}

\begin{figure}
\centering
\includegraphics[height=4.5cm]{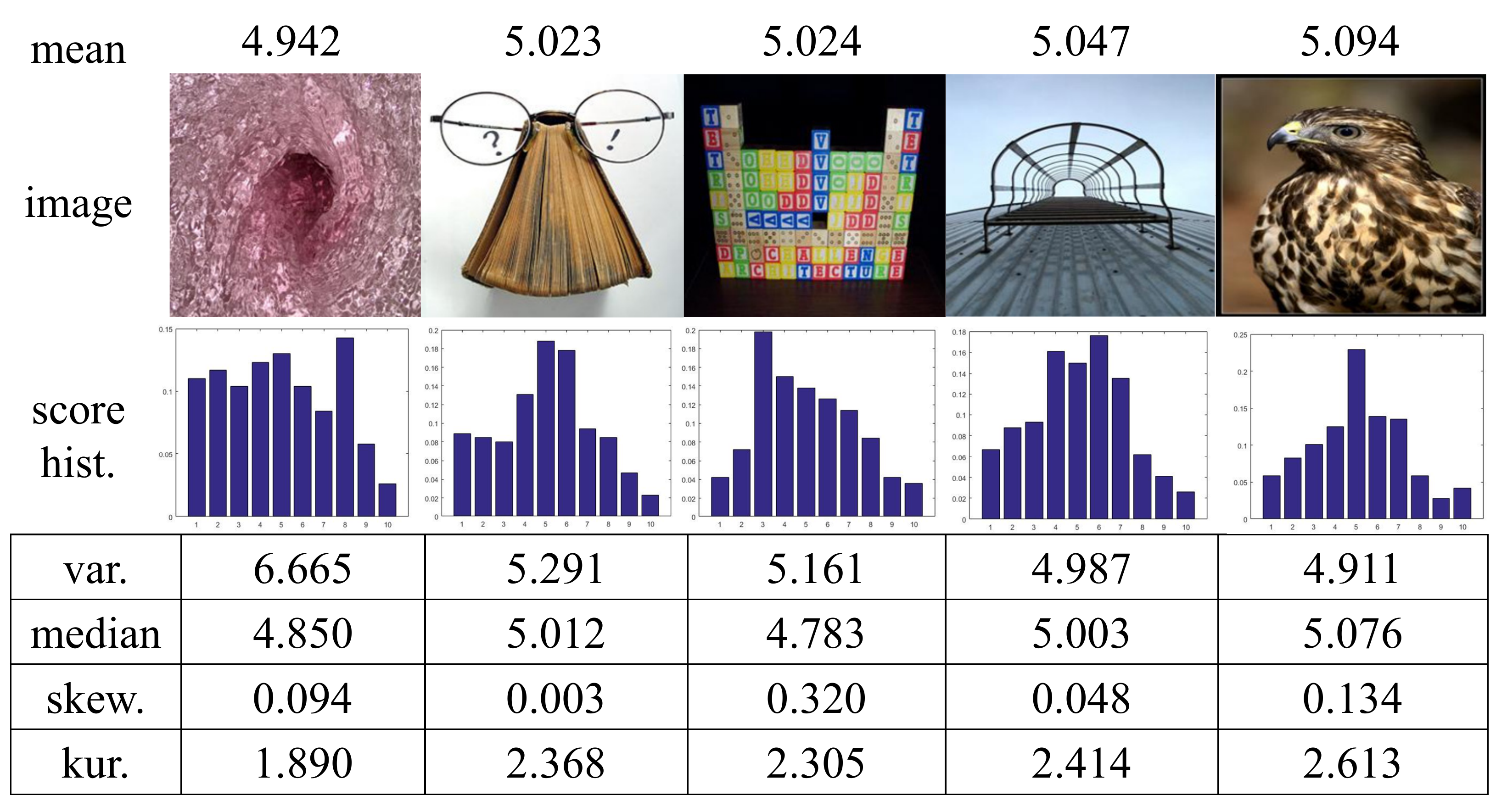}
\caption{Images with similar mean scores (i.e., around 5). The rating distributions are approximated by the score histograms (1-10). The hist., var., skew. and kur. are short for histogram, variance, skewness and kurtosis. The mean scores of the histogram are nearly the same. However, the histograms themselves with their statistics differ from each other. Images are from the AVA dataset \cite{MurrayCVPR2012}, which contains a list of photo IDs from www.dpchallenge.com.}
\label{fig:example}
\end{figure}

Recently, the ability of recognizing the semantic meaning of the objects in an image by computers is greatly increasing through deep convolutional  neural networks. However, recognizing or assessing the aesthetic quality of an image by computers has not reached the practical precision people need.

Subjective Image Aesthetic Quality Assessment (IAQA) is still challenging \cite{MaiCVPR2016} since the large intra class difference of images with high or low aesthetic quality, the large amount of low or high level aesthetic features, and the subjective evaluation of human rating. IAQA has been a hot topic in the communities of Computer Vision (CV), Computational Aesthetics (CA) and Computational Photography (CP).

\textbf{Related work}. As summarized by \cite{DengSPM2017}, in early work, various hand-crafted aesthetic features (i.e. aesthetic rule based features) are designed and connected with a machine classification or regression method. Another line is to use generic image description features. After that, the powerful deep feature representation learned from large amount of data has shown an ever-increased performance on this task, surpassing the capability of conventional hand-crafted features.
\cite{KarayevBMVC2014,LuMM2014,KaoICIP2015,LuICCV2015,LuTMM2015,DongNC2015,KaoSPIC2016,WangSP2016,MaiCVPR2016,KongECCV2016,JinWCSP2016,KaoTIP2017,MaCVPR2017}.


The training data of aesthetic quality assessment are often collected from the online photo sharing communities such as photo.net and dpchallenge.com, in which people rate an image by selecting one of the predefined ordinal basic integer ratings (i.e., 1-7 or 1-10). Higher values indicate better rating \cite{WuICCV2011}. Most of the above studies use the following strategies to encode the aesthetic quality, namely, 1D numerical encoding and binary encoding.

\begin{itemize}

\item \textbf{1D numerical encoding}: the 1-dimension numerical encoding use the weighted mean scores of human ratings. A regression model can be learned to predict the numerical aesthetic quality.

\item  \textbf{binary encoding}: the binary encoding is used to classify the images into high or low aesthetic quality, which is determined by a threshold of the weighted mean scores of human ratings. A classifier can be learned to predict the high-low classification results.

\end{itemize}

However, although there exits consensus of the assessment of image aesthetic quality, it is still a subjective task in nature. The rated scores of multiple persons may differ greatly from each other. People tend to assign inconsistent scores to the same image \cite{WuICIP2010}. There is ambiguity in the image aesthetic quality assessment \cite{KeCVPR2006}. A scalar value is insufficient to capture the true nature of the subjectivity of image aesthetic quality \cite{WuICCV2011}. The main limitation of the above representations is that they do not provide an indicator of the degree of consensus or diversity of opinion among annotators \cite{MurrayCVPR2012}.

Figure \ref{fig:example} shows some images from the AVA dataset \cite{MurrayCVPR2012}. Images with nearly the same mean scores (i.e., around 5) are listed. However, the distributions (approximated by the score histogram) are not that similar. Other statistics such as the variance, the median, the skewness, and the kurtosis differ greatly from each other. The human ratings are quite subjective. The mean score is greatly influenced by the low and high extremes of the rating scale, which makes it inappropriate to be a robust estimation of the whole distribution, especially when the distribution is skewed. For skewed distributions, the median value appears to be more appropriate to describe the distributions than the mean value \cite{WuICCV2011}. The Gaussian distribution is the best-performing model for only 62\% of images in AVA \cite{MurrayCVPR2012}. The others are the skewed ones and can be best fitted by the Gamma distribution \cite{MurrayCVPR2012}.

Most recently, some methods are proposed to use modified or generated score distributions for binary classification and numerical assessment on aesthetics \cite{JinICIP2016,WangIJCNN2017,HouArXiv2016}. Wu et al. \cite{WuICCV2011} propose a modified support vector regression algorithm to predict the score distribution in two small aesthetic datasets, before the large scale AVA dataset released and the popularity of deep CNNs. 

Jin et al. \cite{JinICIP2016} use the weighted Chi-square distance as the loss function to predict the mean score and the standard deviation from the score distribution. Wang et al. \cite{WangIJCNN2017} explicitly modify the score distribution of the AVA dataset as Gaussian and jointly predict its mean and standard deviation. They use the asymmetrical  Kullback-Leibler (KL) divergence as the loss function for their DBN network. Hou et al. \cite{HouArXiv2016} generate score distribution by mapping the real number labels to 10 aesthetic bins of the AADB dataset \cite{KongECCV2016}. They propose to use squared Earth mover's distance (EMD) as the loss function, which can be equivalent to the Euclidean distance of the two cumulative distribution functions for the ordinal basic human ratings prediction. Thus, the loss functions of \cite{HouArXiv2016} and \cite{WuICCV2011} are the same. Note that, all these methods use modified or generated score distributions for binary classification and numerical assessment on aesthetics. While our work is to directly predict the score distribution itself. Murray et al. \cite{MurrayArXiv2017} use the Huber loss combined with ResNet and SPPNet to predict the aesthetic score distribution of an image. Cui et al. \cite{CuiSigIR2017} propose to use the traditional LDL (Label Distribution Learning) technology to predict the aesthetic score distribution of an image.

\textbf{Our Approach}. In this work, we learn from the large aesthetic dataset to predict the aesthetic score distribution of an image, which is represented as a score vector (histogram) using the deep convolutional neural network (DCNN), so as to better capture the subjectivity of aesthetic quality assessment. Conventional CNN which aims to minimize the difference between the predicted scalar numbers or 0-1 classification vectors and the ground truth cannot be directly used for the ordinal basic rating distribution. Inspired by recent work on non-parametric Jensen-Shannon Divergence by Nguyen et al. \cite{NguyenECML2015}, a Cumulative distribution with Jensen-Shannon divergence based CNN (CJS-CNN) is presented to predict the aesthetic score distribution of human ratings. In addition, to alleviate the problem of unreliable human ratings, we propose a new reliability-sensitive learning method based on the kurtosis of the score distribution. The proposed kurtosis can be directly computed using the normalized score histogram. While the rating number is additional information of the normalized score histogram and is not always available in the training set. We compare the recently proposed loss functions designed for score distribution and LDL method with our CJS loss and RS-CJS loss in the experiments. Experimental results on large scale aesthetic dataset demonstrate the effectiveness of our introduced CJS-CNN in this task. The main contributions of our work can be summarized as follows:

\begin{itemize}

\item The first work that predicts a score distribution vector of the ordinal basic human ratings under the deep convolutional neural network framework on the large scale AVA dataset, which is designed to capture the subjectiveness of the human aesthetic quality assessment. 

\item A novel CNN called the CJS-CNN (Cumulative distribution function with Jensen-Shannon Divergence) is introduced. Extensive comparisons with probability distribution function with Euclidean distance, cross entropy distance, Jensen-Shannon Divergence and cumulative distribution function with Euclidean distance are presented.

\item A new reliability-sensitive learning method is proposed based on the kurtosis of the score distribution.


\end{itemize}

Besides the overall aesthetic quality of an image, there are other targets related to aesthetics. Our score distribution prediction can be used for aesthetic image retrieval, guide for shooting good photos, automatic selector for the most aesthetic or attractive cover of a video, etc. From the score distribution, rich information can be outputted, such as mean, median, variance, skewness and kurtosis. For skewed distributions, the median value appears to be more appropriate to describe the distributions than the mean value. The variance, skewness and kurtosis can be jointly used to measure the controversy of an image. The controversy is a measure of the degree of consensus and diversity of the aesthetic assessment of an image. Some artworks may not be accepted today, but may yield potential fashion or masterpieces in the future.

\section{Subjectiveness Analysis of the AVA Dataset}
The assessment of image aesthetic quality is subjective in nature. The perception of aesthetics is affected by the nationality, ethnicity, era, age, education, emotion and many other factors of human beings. In this section we make a statistical analysis of subjectiveness or diversity of the opinion among annotators in a large-scale database for aesthetic visual analysis (AVA) \cite{MurrayCVPR2012}. This dataset is specifically constructed for the purpose of learning more about image aesthetics. All those images are directly downloaded from dpchallenge.com. For each image in AVA, there is an associated distribution of scores (1-10) voted by different viewers. The number of votes that per image gets is ranged in 78-549, with an average of 210, which enables us to have a deeper understanding of such distributions and deduce more information from them.

\begin{figure}
\centering
\includegraphics[height=4cm]{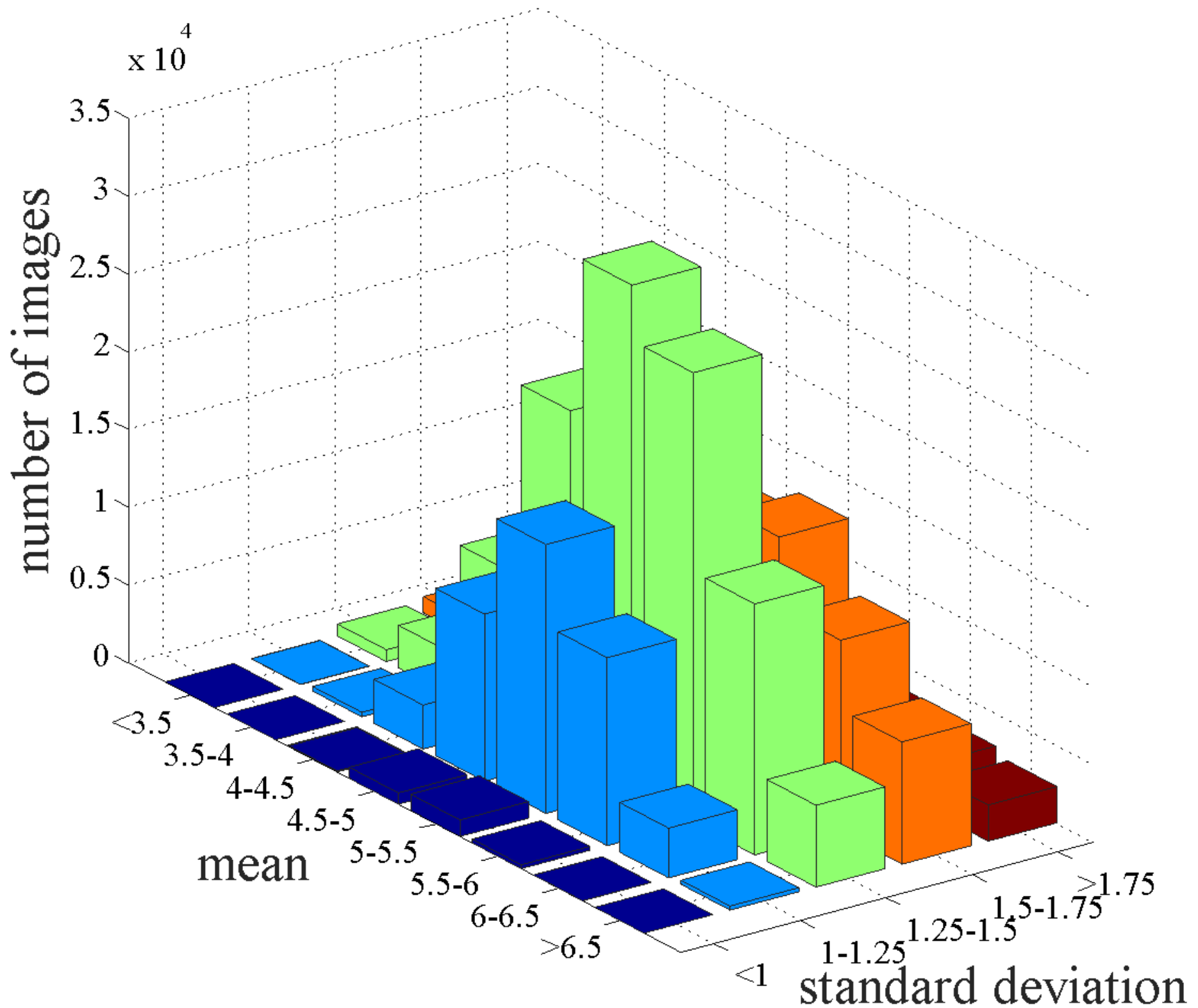}
\caption{The histogram of numbers of images located in different
intervals of the mean and standard deviation of the AVA dataset \cite{MurrayCVPR2012}. }
\label{fig:std}
\end{figure}

\textbf{The Standard Deviation or Variance}. 
As described above, a mean score or a binary high-low label reveals only part of the information deduced from a score distribution. We make a statistical analysis on the number of images according to mean and standard deviation of the human ratings. The standard deviation represents the degree of consensus or diversity of human ratings for the same image, with a higher value meaning higher diversity. The number of images located in each mean and standard deviation interval is shown as a 2D histogram in Figure \ref{fig:std}. Most images' mean values are located in $[4,7]$. Images in this interval are not easy to be classified to a high-low label.  Most images' standard deviation values are larger than $1.25$, which shows the diversity of the human ratings for the same image. In addition, as described in \cite{MurrayCVPR2012}, the variance or standard deviation tends to increase with the distance between the mean score and the mid-point of the rating scale.

\begin{figure}
\centering
\includegraphics[height=3cm]{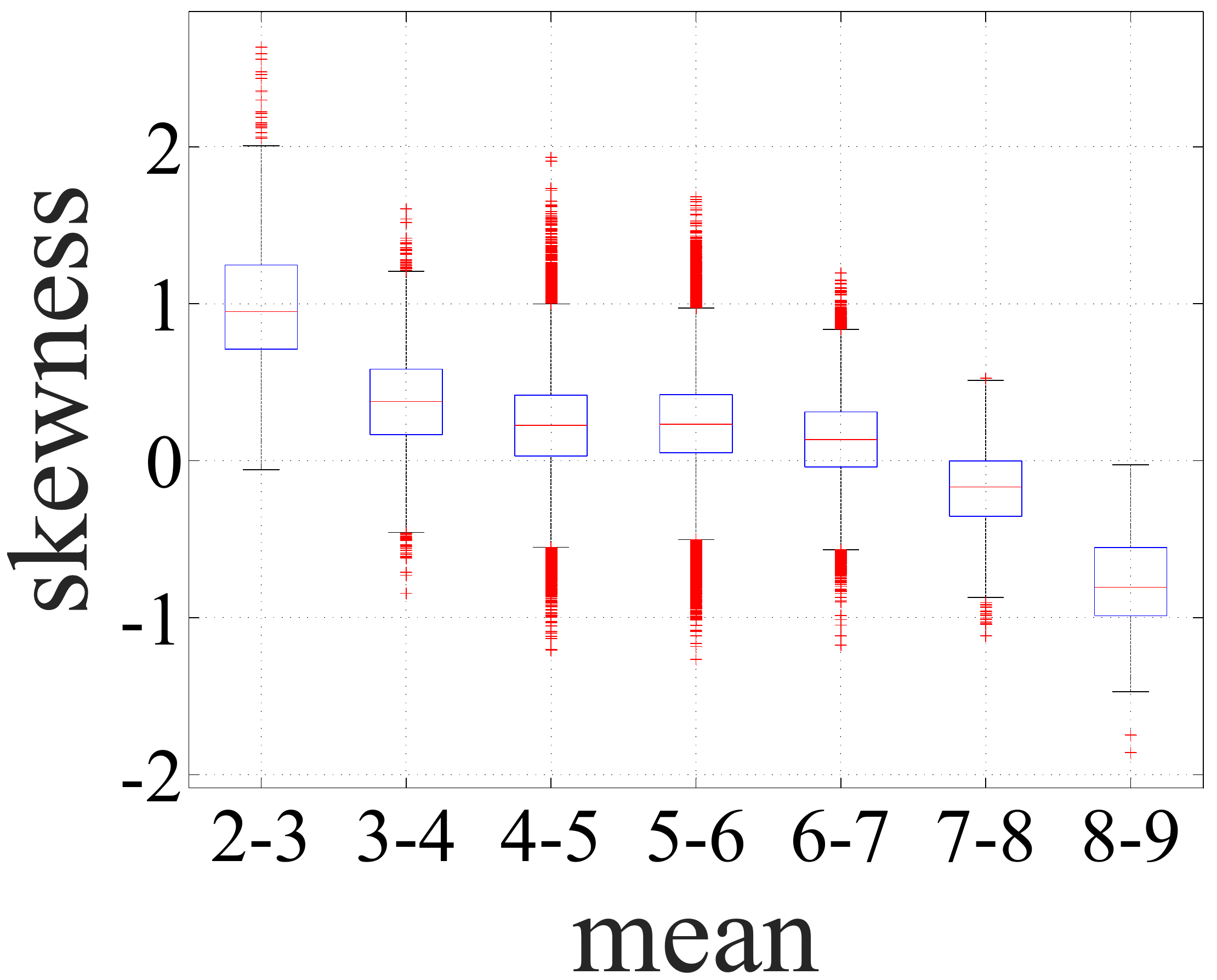}
\includegraphics[height=3cm]{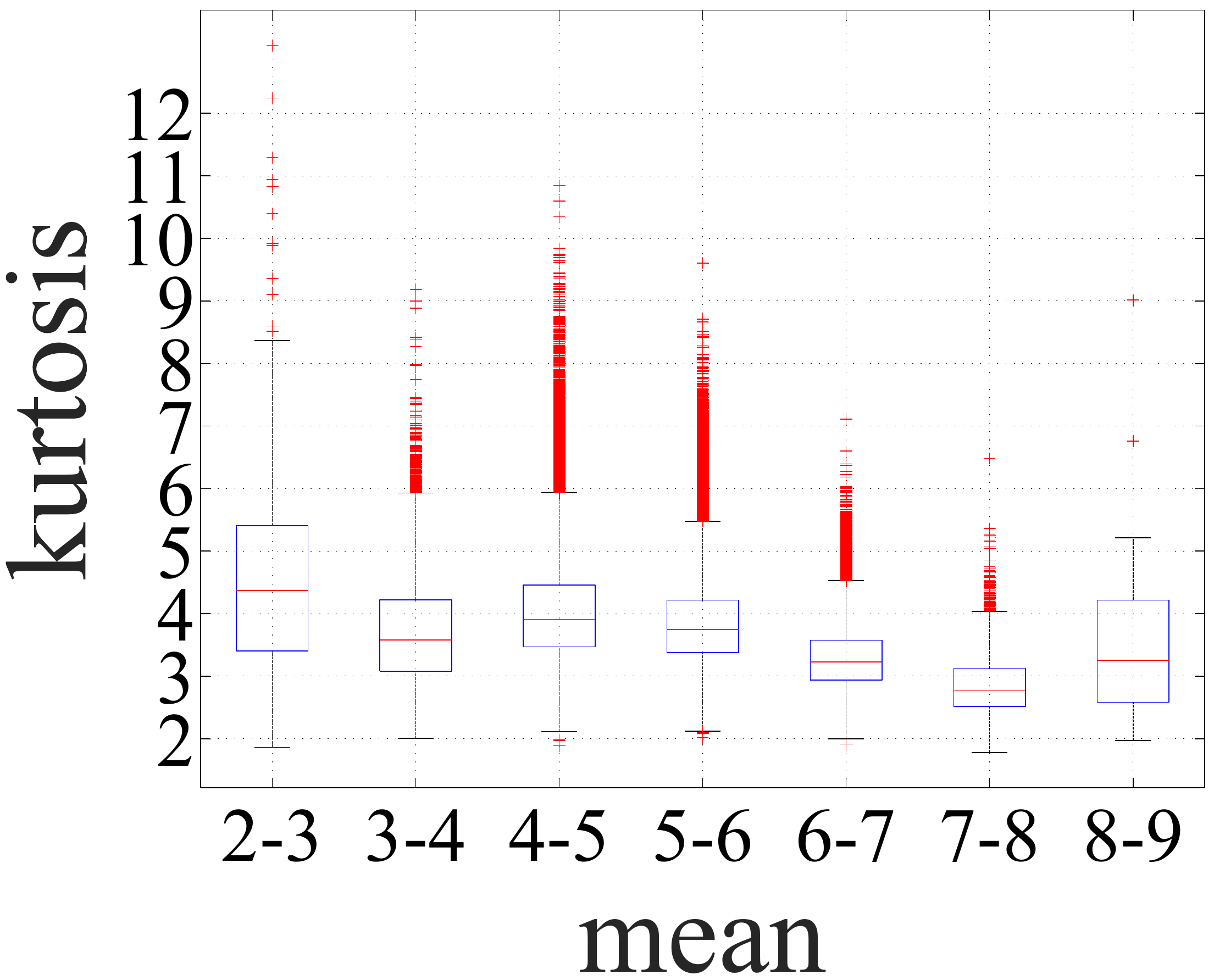}
\caption{Left: Distributions of skewness of score distributions, for images
with different mean scores. The red crosses are the outliers. The skewness tends to decrease from positive to negative with the mean score increasing.
Right: Distributions of kurtosis of score distributions, for images
with different mean scores.}
\label{fig:skew_kur}
\end{figure}

\textbf{The Skewness}.
The skewness \cite{JoanesCMSSK1998,Brown2016} is a measure of the asymmetry of the probability distribution of a real-valued random variable about its mean. If the bulk of the data is at the left and the right tail is longer, we say that the distribution is skewed right or positively skewed; if the peak is toward the right and the left tail is longer, we say that the distribution is skewed left or negatively skewed. Boxplots of the skewness of score distributions for images with mean scores within a specified range are shown in the left side of Figure \ref{fig:skew_kur}. The skewness is a function of mean score in the AVA dataset. Images with mean score values from 4 to 7 tend to have a low absolute value of the skewness and can be considered as those with symmetrical score distributions. Images with mean score values lower than 4 and greater than 7 can be considered as those with positively and negatively skewed score distributions, respectively. This is likely due to the non-Gaussian nature of score distributions at the extremes of the rating scale \cite{MurrayCVPR2012}. 

\begin{figure}
\centering
\includegraphics[height=6.5cm]{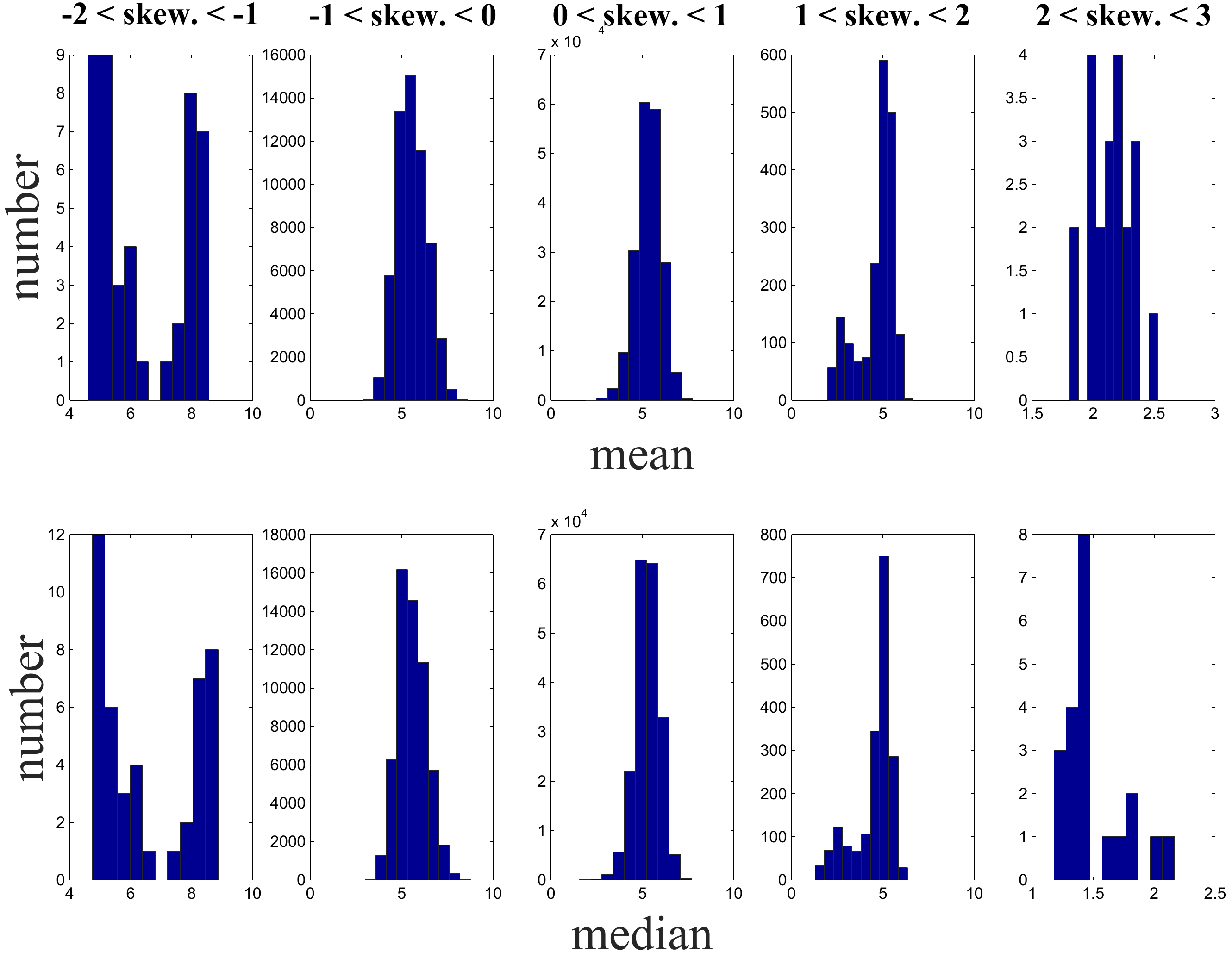}
\caption{Distributions of mean and median of score distributions, for images
with different skewness scores. The divergences between the mean and the median distributions tends to increase with the distance between the skewness values and 0, which is the skewness of the symmetrical normal distribution.}
\label{fig:mean_median}
\end{figure}

Most representative distributions in the AVA dataset are slightly skewed or heavily skewed. For skewed distributions, the median value appears to be more appropriate to describe the distributions than the mean value \cite{WuICCV2011}. The mean and the median values of score distributions for images with skewness within a specified range are shown in Figure \ref{fig:mean_median}. Images with low and high absolute values of the skewness can use the mean and the median to describe their score distributions, respectively.

\textbf{The Kurtosis}.
The other common measure of shape is called the kurtosis \cite{JoanesCMSSK1998,Brown2016}. As skewness is the third moment of the distribution, kurtosis is the fourth moment. The kurtosis of a normal distribuation is 3. A distribution with kurtosis $<3$ and kurtosis $>3$ are called platykurtic and leptokurtic, receptively. Compared with a normal distribution, the platykurtic has shorter and thinner tails and its central peak is lower and broader and vice versa. Score distributions with larger absolute values of the kurtosis (after normalized by minus 3, i.e., normalizing the kurtosis of the normal distribution to 0) have larger divergences from the normal distribution. Boxplots of the kurtosis of score distributions for images with mean scores within a specified range are shown in the right side of Figure \ref{fig:skew_kur}. Within each range of the mean scores, there exist some images with high absolute values of kurtosis values (after normalized by minus 3), which are considered as those with unreliable score distributions.

\section{CJS based CNN for Score Hist. Prediction}
In this section, we introduce the proposed CJS-CNN (Cumulative distribution function with Jensen-Shannon divergence) and the reliability-sensitive learning method based on the kurtosis of the score distribution.

\subsection{The Score Distribution Representation}
With empirical data of the ordinal basic human ratings of an image from the AVA dataset, we use the score histogram to approximate the score distribution. We follow the definition in Wu et al. \cite{WuICCV2011}.

Assuming that there are $Z$ ordinal basic ratings $R = \{R_1,...R_Z\}$. In the AVA dataset, $Z=10, R = \{R_1,...R_{10}\}$. The human ratings for an image can be represented as $S = \{S(1),...,S(L)\}$, where $S(i) \in R$ is given by the $i^{th}$ person and $L$ is the number of persons who have rated this image. (In the AVA dataset, $L \in [78,549]$, with an average of 210). Then the score histogram or score vector of an image in the AVA dataset can be defined as:

\begin{eqnarray}
\begin{split}
y = \{h(1),...,h(i),...,h(Z)\}\\
h(i) = \frac{\sum_{j}^{L}\delta(S(j)=R_i)}{L},
\end{split}
\label{eq:hist}
\end{eqnarray}
where $\delta()$ is the indication function. With this representation, we can calculate the mean, median, variance, skewness, kurtosis using textbook methods.

\subsection{The CJS-CNN}
\label{sec:CJS-CNN}

We use the first $1/3$ part of the GoogLeNet (layers before the first softmax layer) as our DCNN for fast training and extensive comparisons. We replace the full connected layer before the first softmax layer of the GoogLeNet with a output layer of $Z=10$ dimensions. After each element of the output layer, we add a sigmoid layer to normalize each element to $[0,1]$. The layers after the first softmax layer of the GoogLeNet are removed for fast training and comparisons.

The score vector defined by Eq. \ref{eq:hist} can be considered as a vector. A straightforward way to calculate the loss is using the Euclidean distance. However, the score vector is an approximate of the underline probability distribution function (pdf). In addition, the score vector is built on the pre-defined ordinal basic ratings.  Thus, a divergence between two cumulative distribution functions (cdf) is more appropriate  for the loss function. Recently, Nguyen et al. \cite{NguyenECML2015} propose a non-parametric Jensen-Shannon divergence, which performs well in detecting differences between distributions, outperforming the state-of-the-art methods in both statistical power and efficiency for a wide range of tasks. As verified by \cite{NguyenECML2015}, the CJS is quite suit for  non-parametric computation on empirical data without estimating the underline distribution, such as the ordinal basic rating data of the AVA dataset. They define the asymmetrical continuous cumulative Jensen-Shannon divergence ($ACCJS(p(X)||q(X))$) of two continuous probability distribution functions $p(X)$ and $q(X)$ as follows.

\begin{small}
\begin{equation}
\int P(x)log~\frac{P(x)}{\frac{1}{2}P(x)+\frac{1}{2}Q(x)}dx + \frac{1}{ln2} \int (Q(x)-P(x))dx
\end{equation}
\label{eq:ASCJS}
\end{small}

The cumulative distribution function $Y$ of the probability distribution function $y$ defined by Eq. \ref{eq:hist} is defined as follows.

\begin{equation}
Y(i) = \sum_{j=1}^{i} y(j)
\label{eq:cdf}
\end{equation}

\textbf{CJS}. Thus, we define the symmetrical discrete cumulative Jensen-Shannon divergence ($CJS(y_1||y_2)$) of two score histograms $y_1$ and $y_2$ defined by Eq. \ref{eq:hist} as follows, derived from ($ACCJS(p(X)||q(X))+ACCJS(q(X)||p(X))$).

\begin{small}
\begin{equation}
\frac{1}{2}
[
\sum_{i=1}^{Z}Y_1(i)log~\frac{Y_1(i)}{\frac{1}{2}Y_1(i)+\frac{1}{2}Y_2(i)} 
+ 
\sum_{i=1}^{Z}Y_2(i)log~\frac{Y_2(i)}{\frac{1}{2}Y_1(i)+\frac{1}{2}Y_2(i)}
], 
\end{equation}
\end{small}
\label{eq:D-CJS}
where $Y_1$ and $Y_2$ are defined by Eq. \ref{eq:cdf}. After that, we define our \textbf{CJS} loss function for the CJS-CNN as:

\begin{equation}
l^{CJS}(y,\hat{y}) = CJS(y||\hat{y}),
\label{eq:CJS}
\end{equation}
where $y$ is the ground truth score histogram, and $\hat{y}$ is the predicted score histogram by our CJS-CNN.

\subsection{The Reliability-sensitive Learning}
In Eq. \ref{eq:hist}, the larger the rating number $L$ is, the more reliable the distribution is. Wu et al. \cite{WuICCV2011} use the rating numbers to model the reliability of the score distribution. In the AVA dataset, the number of votes that per image gets is ranged in 78-549 with an average of 210, which limits the performance of the rating number based reliability learning. Besides, one cannot obtain the rating numbers from normalized score histograms. If another dataset has only normalized score histograms, one cannot use the rating number for the reliability learning.

\textbf{RS-CJS}. We propose to use the kurtosis to measure the reliability of a score distribution $y$ defined by Eq. \ref{eq:hist}. The kurtosis of a normal distribution is 3. Score distributions with kurtosis closer to 3 have smaller divergence from the normal distribution. Thus, inspired by Wu et al. \cite{WuICCV2011}, we define the reliability factor $r^{kurtosis}$ as follows.

\begin{eqnarray}
\begin{split}
r^{kurtosis}(y) & = \mu(T(y)), T(y) = \frac{1}{|kus(y)-3|}\\
\mu(T(y)) & = 
\begin{cases}
  \frac{ln(T(y)+1)}{ln(T(y)+1)+1}, & T(y)<Th  \\ 
  1,  & \text{otherwise}  
\end{cases},
\label{eq:r_kus}
\end{split}
\end{eqnarray}
where $r^{kurtosis}(y)$ equals to 1 if the kurtosis $kus(y)$ is sufficiently close to 3 and tends to 0 if $|kus(y)-3|$ is very large. In practice, we add a small number $\epsilon$ to $|kus(y)-3|$ to avoid the division by zero. We choose the threshold $Th$ using cross validation. 
In practice, we use the percentage of the number of images above $Th$ against the total number of the training images to determine $Th$. When the percentage equals $10\%$, we obtain the best performance of our score distribution prediction task (the other candidate percentages are: $5\%, 20\%, 30\%$).



%

Thus, the reliability-sensitive CJS loss is defined as:

\begin{equation}
l^{RS-CJS}(y,\hat{y}) = r^{kurtosis}(y) CJS(y,\hat{y}),
\label{eq:RS-CJS}
\end{equation}
where $y$ is the ground truth score histogram in the AVA dataset, and $\hat{y}$ is the predicted score histogram by our CJS-CNN. The more reliable the training image is, the more penalty it should obtain when the prediction is not correct.


\section{Experiments}
\begin{figure}
\centering
\includegraphics[width=8.3cm]{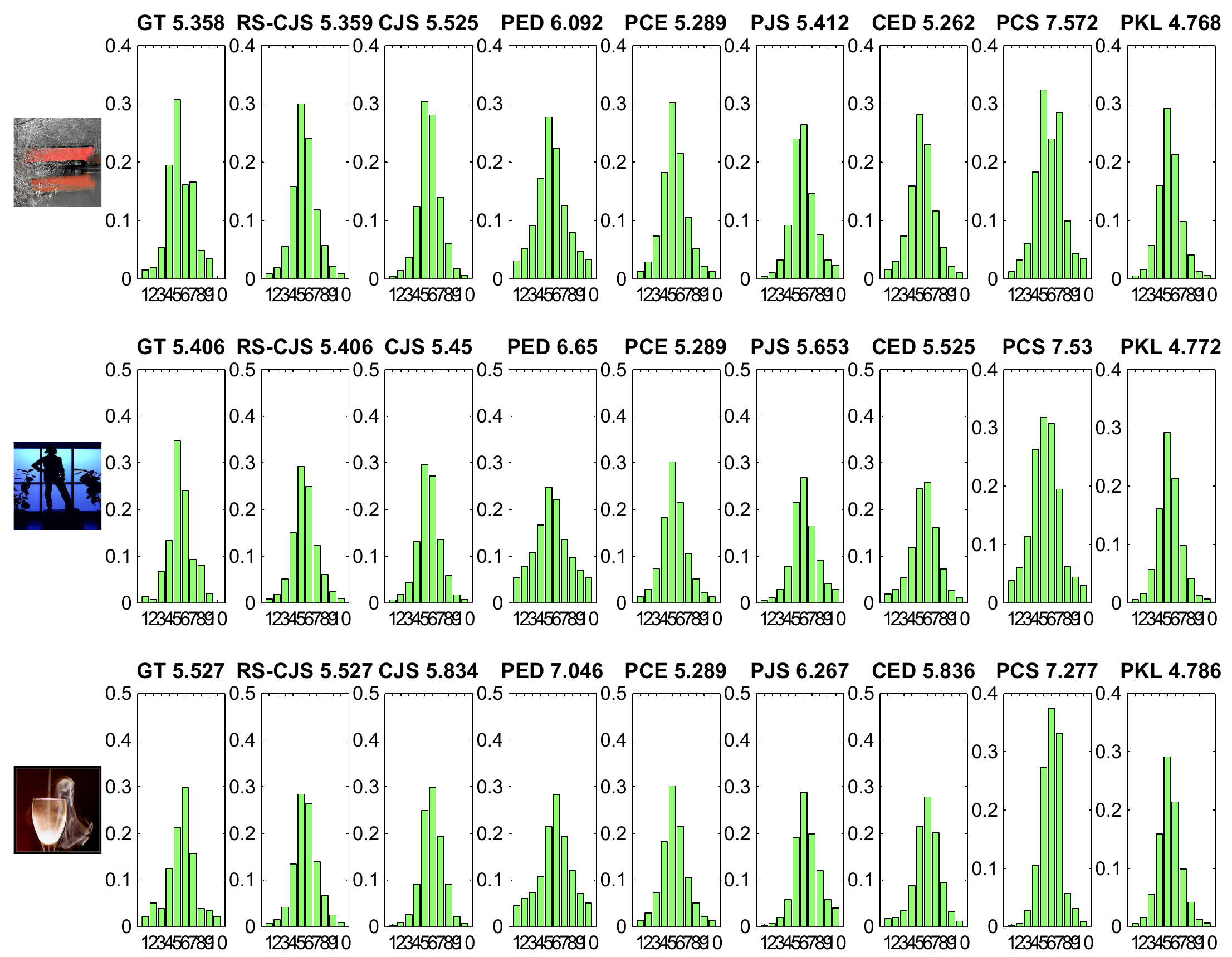}
\caption{Predicted score histograms by the above loss functions. The numbers above each histograms are their mean scores. The first column is the images. The 2nd column is the human rating distributions (GT: Ground Truth). The 3rd and the 4th columns are the results predicted by our proposed RS-CJS and CJS based CNN, respectively. The other columns are the predicted results of other loss functions. Our results are more similar to the ground truth of human ratings than others. Images are from the AVA dataset \cite{MurrayCVPR2012}, which contains a list of photo IDs from www.dpchallenge.com.}
\label{fig:hist}
\end{figure}

In this section, we present the experimental results in the AVA dataset. We follow the standard partition method of the AVA dataset in previous work
\cite{MurrayCVPR2012,WangSP2016,KongECCV2016,LuTMM2015,LuICCV2015,MaiCVPR2016}
. The training and test sets contain 235,599 and 19,930 images respectively. In all the experiments, for fair comparisons of various loss functions, we use the first $1/3$ part of the GoogLeNet as the DCNN. We show the predicted score histograms by our proposed CJS-CNN and other compared loss functions on the test set of AVA in Figure \ref{fig:hist}. Our CJS-CNN achieves the most similar results to the ground truth human rating distributions.

\subsection{Implementation Details}


We fix the  parameters of the layers before the first full connected layer of a pre-trained GoogLeNet model \footnote{\url{http://vision.princeton.edu/pvt/GoogLeNet/ImageNet/}} on the ImageNet \cite{DengCVPR2009} and fine tune the 2 full connected layers on the training set of the AVA dataset. We use the Caffe framework \cite{JiaMM2017} to train and test our models. The learning policy is set to \emph{step}. Stochastic gradient descent is used to train our model with a mini-batch size of 48 images, a momentum of 0.9, a gamma of 0.5 and a weight decay of 0.0005. The max number of iterations is 480000. The training time is about 3 days using GTX980-Ti GPU and about 2 days using Titan X Pascal GPU.

\subsection{Score Histogram Prediction and Comparisons}
\subsubsection{Baseline Loss Functions}
\label{sec:baseline}
Besides the CJS loss function we proposed, we also evaluate other distribution divergences based loss functions as the baseline methods, parts of which are described below. These divergences or distances are often used in computer vision and pattern recognition tasks to compute the difference between two distributions or feature vectors.  All the probability or cumulative distribution functions in our paper refer to discrete histograms. The DCNN cooperated with each divergence or distance based loss function is the first $1/3$ part of the GoogLeNet  for fair comparisons.


\begin{table*}[!t] 
\renewcommand{\arraystretch}{1}
\caption{The mean divergences (MD, Eq. \ref{eq:MD}) between the predicted score histogram and the ground truth of various loss functions. The dataset is AVA. In all the methods listed below, the DCNN is the first 1/3 part of the GoogLeNet. The LDL method proposed by  \cite{CuiSigIR2017} does not use DCNN. Except the PED divergences, the performances of the other divergences we use are not reported in their work \cite{CuiSigIR2017}.}
\label{tb:divergences}
\centering
\rowcolors{2}{gray!50}{white}
\begin{tabular}{|c||c|c|c|c|c|c|c|}
\hline
\backslashbox{loss}{MD} & PED & PCE & PJS & PCS & PKL & CED & CJS \\
\hline\hline
PED & 0.197 & 2.830 & 0.059 & 0.105 & 0.728 & 0.323 & 0.068\\
RS-PED & 0.189 & 2.733 & 0.055 & 0.094 & 0.657 & 0.324 & 0.067\\
\hline
PCE & 0.167 & 2.773 & 0.041 & 0.075 & 0.442 & 0.279 & 0.049\\
RS-PCE & 0.169 & 2.771 & 0.046 & 0.071 & 0.438 & 0.279 & 0.047\\
\hline
PJS & 0.185 & 2.828 & 0.051 & 0.093 & 0.527 & 0.326 & 0.053\\
RS-PJS & 0.183 & 2.776 & 0.049 & 0.091 & 0.523 & 0.327 & 0.049\\
\hline
PCS \cite{JinICIP2016} & 0.182 & 2.807 & 0.045 & 0.082 & 0.450 & 0.287 & 0.045\\ 
RS-PCS & 0.175 & 2.783 & 0.045 & 0.079 & 0.423 & 0.277 & 0.044\\ 
\hline
PKL \cite{WangIJCNN2017} & 0.163 & 2.779 & 0.039 & 0.073 & 0.389 & 0.270 & 0.044\\ 
RS-PKL & 0.164 & 2.778 & 0.037 & 0.071 & 0.386 & 0.268 & 0.043\\ 
\hline
MMD \cite{BorgwardtISMB2006} & 0.201 & 2.831 & 0.064 & 0.112 & 0.710 & 0.339 & 0.068\\ 
RS-MMD & 0.196 & 2.824 & 0.063 & 0.097 & 0.710 & 0.322 & 0.054\\ 
\hline
Huber \cite{MurrayArXiv2017} & 0.184 & 2.775 & 0.044 & 0.078 & 0.409
 & 0.279 & 0.053\\ 
RS-Huber & 0.183 & 2.774 & 0.045 & 0.074 & 0.402 & 0.271 & 0.048\\ 
\hline
CED \cite{WuICCV2011,HouArXiv2016} & 0.182 & 2.799 & 0.047 & 0.085 & 0.502 & 0.294 & 0.049\\  
RS-CED & 0.180 & 2.792 & 0.048 & 0.082 & 0.502 & 0.283 & 0.047\\  
\hline
\textbf{Our CJS} & 0.163 & 2.779 & 0.039 & 0.072 & 0.382 & 0.266 & 0.041\\
\textbf{Our RS-CJS} & \textbf{0.158} & \textbf{2.760} & \textbf{0.037} & \textbf{0.068} & \textbf{0.381} & \textbf{0.260} & \textbf{0.040}\\
\hline
LDL Method \cite{CuiSigIR2017} & 0.303 & - & - & - & - & - & - \\
\hline
\end{tabular}
\end{table*}

\begin{table*}[!t] 
\renewcommand{\arraystretch}{1}
\caption{The ablation study of $\lambda$ in Eq. \ref{eq:lambda}. The dataset is AVA. The DCNN is the first 1/3 part of the GoogLeNet.} 
\label{tb:lambda}
\centering
\begin{tabular}{|c||c|c|c|c|c|c|c|}
\hline
\backslashbox{loss}{MD} & PED & PCE & PJS & PCS & PKL & CED & CJS\\
\hline\hline
$\lambda = 0$ & 0.159 & 2.760 & 0.037 & 0.068 & 0.387 & 0.260 & 0.040 \\
\hline
$\lambda = 0.1$ & 0.159 & 2.764 & 0.038 & 0.069 & 0.384 & 0.262 & 0.040\\
\hline
$\lambda = 0.3$ & 0.159 & 2.762 & 0.038 & 0.069 & 0.386 & 0.262 & 0.040\\
\hline
$\lambda = 0.5$ & 0.160 & 2.766 & 0.038 & 0.070 & 0.386 & 0.264 & 0.040\\
\hline
$\lambda = 0.7$ & 0.158 & 2.761 & 0.037 & 0.068 & 0.385 & 0.261 & 0.041\\ 
\hline
$\lambda = 0.9$ & 0.159 & 2.763 & 0.038 & 0.069 & 0.384 & 0.262 & 0.040\\  
\hline
\textbf{Our RS-CJS ($\lambda = 1$)} & \textbf{0.158} & \textbf{2.760} & \textbf{0.037} & \textbf{0.068} & \textbf{0.381} & \textbf{0.260} & \textbf{0.040}\\
\hline
\end{tabular}
\end{table*}

\textbf{PED}. The loss function using the Euclidean distance of the two probability distribution functions is defined as:

\begin{equation}
l^{PED}(y,\hat{y}) = \sum_{i=1}^{Z}(y(i)-\hat{y}(i))^2
\label{eq:PED}
\end{equation}

\textbf{PCE}. The loss function using the cross entropy of the two probability distribution functions is defined as:

\begin{equation}
l^{PCE}(y,\hat{y}) = -\sum_{i=1}^{Z}[(y(i)log~\hat{y}(i)+(1-y(i))log~(1-\hat{y}(i))]
\label{eq:PCE}
\end{equation}
This is the standard and widely used loss function for image classification problems and can be used as histogram difference for our task.

\textbf{PJS}. The loss function using the symmetrical version of the Jensen-Shannon divergence of the two probability distribution functions is defined as:

\begin{small}
\begin{equation}
l^{PJS}(y,\hat{y}) = 
\frac{1}{2}
[
\sum_{i=1}^{Z}y(i)log~\frac{y(i)}{m(y,\hat{y})} 
+ 
\sum_{i=1}^{Z}\hat{y}(i)log~\frac{\hat{y}(i)}{m(y,\hat{y})}
],
\label{eq:PJS}
\end{equation}
\end{small}
where $m(y,\hat{y}) = \frac{1}{2}y(i)+\frac{1}{2}\hat{y}(i)$.

\textbf{PCS} \cite{JinICIP2016}. The loss function using the Chi-square distance of the two probability distribution functions is defined as:

\begin{small}
\begin{equation}
l^{PCS}(y,\hat{y}) = 
\frac{1}{2}
\sum_{i=1}^{Z}\frac{(y(i)-\hat{y}(i))^2}{y(i)+\hat{y}(i)}
\label{eq:PCS}
\end{equation}
\end{small}
This loss function is proposed by Jin et al. \cite{JinICIP2016} to predict the mean score and standard deviation from the score distribution.

\textbf{PKL} \cite{WangIJCNN2017}. The loss function using the symmetrical version of the Kullback–Leibler divergence of the two probability distribution functions is defined as:

\begin{small}
\begin{equation}
l^{PKL}(y,\hat{y}) = 
\frac{1}{2}
[
\sum_{i=1}^{Z}y(i)log~\frac{y(i)}{\hat{y}(i)} 
+ 
\sum_{i=1}^{Z}\hat{y}(i)log~\frac{\hat{y}(i)}{y(i)}
]
\label{eq:PKL}
\end{equation}
\end{small}
The asymmetrical version of the KLD loss is used by Wang et al. \cite{WangIJCNN2017}, who 
explicitly modify the score distribution of the AVA dataset as Gaussian and jointly predict its mean and standard deviation.

\textbf{CED} \cite{WuICCV2011,HouArXiv2016}. The loss function using the Euclidean distance of the two cumulative distribution functions is defined as:

\begin{equation}
l^{CED}(y,\hat{y}) = \sum_{i=1}^{Z}(Y(i)-\hat{Y}(i))^2,
\label{eq:CED}
\end{equation}
where $Y$ and $\hat{Y}$ are the cumulative distribution functions of the original probability distribution functions $y$ and $\hat{y}$, as defined in Eq. \ref{eq:cdf}. This loss function is also used in Wu et al. \cite{WuICCV2011} and can be derived from the squared Earth mover's distance (EMD) by Hou et al.  \cite{HouArXiv2016} for the ordinal basic human ratings prediction.

%

We show the predicted score histograms by our proposed CJS-CNN and other compared loss functions on the test set of AVA in the supplementary material. Our CJS-CNN achieves the most similar results to the ground truth human rating distributions.

%
%

%

\subsubsection{Numerical Evaluation Results} 
%
%

In Table \ref{tb:divergences}, we summarize the evaluation results of the loss functions over the divergences. We use the Mean Divergences (MD) to evaluate various divergences between the predicted score histogram and the ground truth on the test set of AVA. The MD is defined as: 


%
\begin{equation}
\text{MD} = \frac{1}{n}\sum_{i=1}^{n}l(y,\hat{y}),
\label{eq:MD}
\end{equation}
where $l=\{l^{PED},l^{PCE},l^{PJS},l^{PCS},l^{PKL},l^{CED},l^{CJS}\}$ defined above. $n$ is size of the test set.


The results in Table \ref{tb:divergences} reveal that, our proposed RS-CJS and CJS based CNN outperform other methods. Among all the odd lines with white background, our CJS achieves the best performance. All the mean divergences of our RS-CJS on the test sets are the smallest. Typically, in a learning setting, optimizing directly a certain criterion should lead to higher performance than optimizing a related one. However, although our methods are optimizing the CJS loss, the learned model can achieve best performance in other related loss. This is mainly because that, as verified by \cite{NguyenECML2015}, the CJS is quite suitable for non-parametric computation on empirical data, such as the ordinal basic rating data of the AVA dataset. The line with header 'RS-' means adding our reliability-sensitive learning strategy. Almost all the RS version methods (the even lines) perform better than the corresponding ones (the odd lines). The reliability sensitive learning based on the kurtosis reduces the impacts of the unreliable training samples.

\subsubsection{The Ablation Study of the Reliability Factor}
Wu et al. \cite{WuICCV2011} propose to use the number of ratings of each image for the reliability factor $r^{ratnum}(y)$. The larger the rating number is, the larger the reliability of rating is. To compare with our kurtosis based reliability factor $r^{kurtosis}(y)$ in Eq. \ref{eq:r_kus} and Eq. \ref{eq:RS-CJS}, we use an balance factor $\lambda$ as follow to make ablation study.

\begin{equation}
r(y) = \lambda r^{kurtosis}(y) + (1-\lambda) r^{ratnum}(y)
\label{eq:lambda}
\end{equation}
For a fair comparison, we use $r(y)$ on the CJS loss: $r(y) CJS(y,\hat{y})$

The comparison results are shown in Table \ref{tb:lambda}. The results reveal that the performance of $r^{kurtosis}(y)$ are slightly better than that of $r^{ratnum}(y)$. The combination of these two reliability factors does not produce better performance. Note that, the kurtosis can be directly computed using the normalized score histogram. While the rating number is additional information of the normalized score histogram and is not always available in the training set.

\section{Conclusions and Discussions}
In this paper, we propose the CJS-CNN to predict the aesthetic score distribution of images. Unlike the object recognition, which definitely has right answers in most cases, the image aesthetic assessment is a subjective task in nature. Thus, only using a scalar to represent the aesthetics may not be the right direction.

Instead of only predicting the binary high-low label or the numerical score, we can output the aesthetic score distribution with rich statistics for various applications such as aesthetic image retrieval, aesthetic image enhancement. The overall aesthetic quality can be represented by the mean or median. The controversy or subjectiveness can be measured by the variance. The popularity of an image can be measured by the rating number of human. However, the rating number cannot be derived from the predicted score histogram. As shown in the experiments, we can use the kurtosis to approximate the popularity instead of the rating number.


The aesthetic quality assessment is a subjective task in nature. It has been a long time that people focused on the scalar representation (1D numerical or binary coding) of aesthetics. Wu et al. \cite{WuICCV2011} pointed out this problem and made an attempt to predict the score distribution. However, it was submerged in rich literatures which aim to rise the classification or regression accuracy of the scalar representation. With the powerful deep representation learning technologies, we think it is the right time to let the aesthetic quality assessment return to it's subjective nature. This paper is a restart of this direction. We hope it can inspire more work in the future, such as (1) mapping more statistics to the subjective evaluation of vast amount of images, (2) designing new large scale aesthetic datasets with unbiased data and specially for subjective assessment of aesthetics, (3) using more powerful and larger DCNNs or other machine learning technologies to make the assessment by computer better match that of human.


\section{ Acknowledgments}
We thank all the reviewers and ACs. This work is partially supported by the National Natural Science Foundation of China (Grant Nos. 61402021, 61401228, 61402463, 61772513), the Science and Technology Project of the State Archives Administrator (Grant No. 2015-B-10), the open funding project of State Key Laboratory of Virtual Reality Technology and Systems, Beihang University (Grant No. BUAA-VR-16KF-09), the Fundamental Research Funds for the Central Universities (Grant No. 3122014C017), the China Postdoctoral Science Foundation (Grant No. 2015M581841), and the Postdoctoral Science Foundation of Jiangsu Province (Grant No. 1501019A).

{\small
\bibliographystyle{aaai}
\bibliography{aaai18}
}

\end{document}